\pdfoutput=1
\documentclass[11pt]{article}

\usepackage{acl}

\usepackage{times}
\usepackage{latexsym}
\usepackage{comment}
\usepackage{booktabs}
\usepackage{amsmath}
\usepackage{kantlipsum}
\usepackage[inline]{enumitem}
\usepackage{tcolorbox}

\usepackage[T1]{fontenc}
\usepackage[utf8]{inputenc}

\usepackage{microtype}
\newcommand*\samethanks[1][\value{footnote}]{\footnotemark[#1]}

\usepackage{inconsolata}

\usepackage{graphicx}

\title{UTSA-NLP at ArchEHR-QA 2025: Improving EHR Question Answering via Self-Consistency Prompting}

\author{Sara Shields-Menard\thanks{These two authors contributed equally to this work.}, Zach Reimers\samethanks, Joshua Gardner,\\\textbf{David Perry, and Anthony Rios}\\
  The University of Texas at San Antonio\\
  \texttt{\{Sara.Shields-Menard2, Zachary.Reimers2,}\\\texttt{Joshua.Gardner, David.Perry\}@my.utsa.edu}\\
  \texttt{\{Anthony.Rios\}@utsa.edu}}

\begin{document}
\maketitle
\begin{abstract}
We describe our system for the ArchEHR-QA Shared Task on answering clinical questions using electronic health records (EHRs). Our approach uses large language models in two steps: first, to find sentences in the EHR relevant to a clinician’s question, and second, to generate a short, citation-supported response based on those sentences. We use few-shot prompting, self-consistency, and thresholding to improve the sentence classification step to decide which sentences are essential. We compare several models and find that a smaller 8B model performs better than a larger 70B model for identifying relevant information. Our results show that accurate sentence selection is critical for generating high-quality responses and that self-consistency with thresholding helps make these decisions more reliable.
\end{abstract}

\section{Introduction}
% The introduction should be structured as follows:

% 1. What is the problem?
The prevalence of electronic health records (EHRs) has prompted increased digital communication between patients and clinicians. Patient portal use has been linked to better health outcomes for those with chronic conditions, especially by improving self-management and treatment~\cite{brands2022patient} and doctor-patient relationships~\cite{carini2021impact}. While patients benefit from EHRs and patient portals, clinicians now face one extra hour of work per day, much outside of work hours, due to the high volume of patient-initiated messages and results~\cite{akbar2021inboxwork}. The result is increased clinician burden and burnout. This work focuses on developing methods that could potentially answer patient portal questions.

% 2. How is the problem currently addressed?
Large language models are in the early stages of adoption in medical systems and have begun to be implemented in clinical decision support (CDS), medical question-answering systems, and medical documentation~\cite{liu2023chatgptcds}. Clinical decision support (CDS) has been used to help reduce clinician burnout by sending rule-based alerts to clinicians and patients based on their EHRs. The clinician either canceled or ignored these alerts due to improper timing or alert fatigue. Large Language Models (LLMs) such as ChatGPT have shown beneficial use to assist the clinician in alert logic~\cite{liu2023chatgpt}. ChatGPT has also been used as a chatbot assistant to respond to patient questions from a social media forum~\cite{ayers2023chatbot}. Overall, studies have found that AI-generated responses are longer than those written by physicians~\cite{ayers2023chatbot, liu2023chatgpt}. Some models are rated as more empathetic and of higher quality~\citep{ayers2023chatbot}, while reviewers note others to have a noticeably artificial tone~\citep{li2023chatdoctor}. Recent efforts such as Med-PaLM and Med-PaLM 2 have advanced long-form medical QA by incorporating improved prompting strategies, fine-tuning on medical datasets, and human-centric evaluation frameworks~\cite{singhal2023large, medpalm2025toward}. These models demonstrate strong performance on USMLE-style questions and improved physician-rated safety and factuality in long-form answers~\cite{pfohl2024toolbox, callahan2021bedside, ayers2023comparing}.

% 3. What are the weaknesses in approaches?
While promising results have been shown using LLMs with Question-Answer queries for specific medical topics like cancer, hepatic disease, and Obstetrics and gynecology, the use of LLMs for patient-specific question-answer responses is limited~\citep{liu2023chatgpt}. Even with a medical chat model that can respond to patient questions using online medical content~\citep{li2023chatdoctor}, the patient’s question is not directly answered with their own EHR information, which may provide critical references to drug interactions, surgery recoveries, or lab results. One limitation of previous work is that the data to train these models is outdated by the time the model is ready to be deployed, so the use of current patient EHRs is essential for maximum clinician and patient benefit. Further, most LLMs used in prior work are not grounded in real-time, patient-specific data, which has been shown to impact accuracy and safety in recent benchmark evaluations~\cite{medpalm2025toward}.

% 4. How do we overcome weaknesses?
The ArchEHR-QA shared task provides a wide range of real patient EHRs with clinician and patient questions to generate concise responses to patient/clinician questions grounded in the patient EHRs~\citep{soni-etal-2025-archehr-qa}. In this study, we designed a prompting pipeline for LLMs to identify and compile relevant patient notes in response to a clinician’s question. We specifically used few-shot prompting combined with self-consistency and thresholding to select the EHR sentences most relevant for response generation.

Our work contributes the following findings to the solution for the ArchEHR-QA: BioNLP at ACL 2025 Shared Task on Grounded Electronic Health Record Question Answering: 
\begin{itemize}
    \item A method for sentence classification using few-shot prompting with self-consistency and thresholding.
    \item An analysis showing that sentence selection quality is the primary driver of overall performance, with errors concentrated in the \textit{supplementary} class.
\end{itemize}

\begin{figure*}
    \centering
    \includegraphics[width=\linewidth]{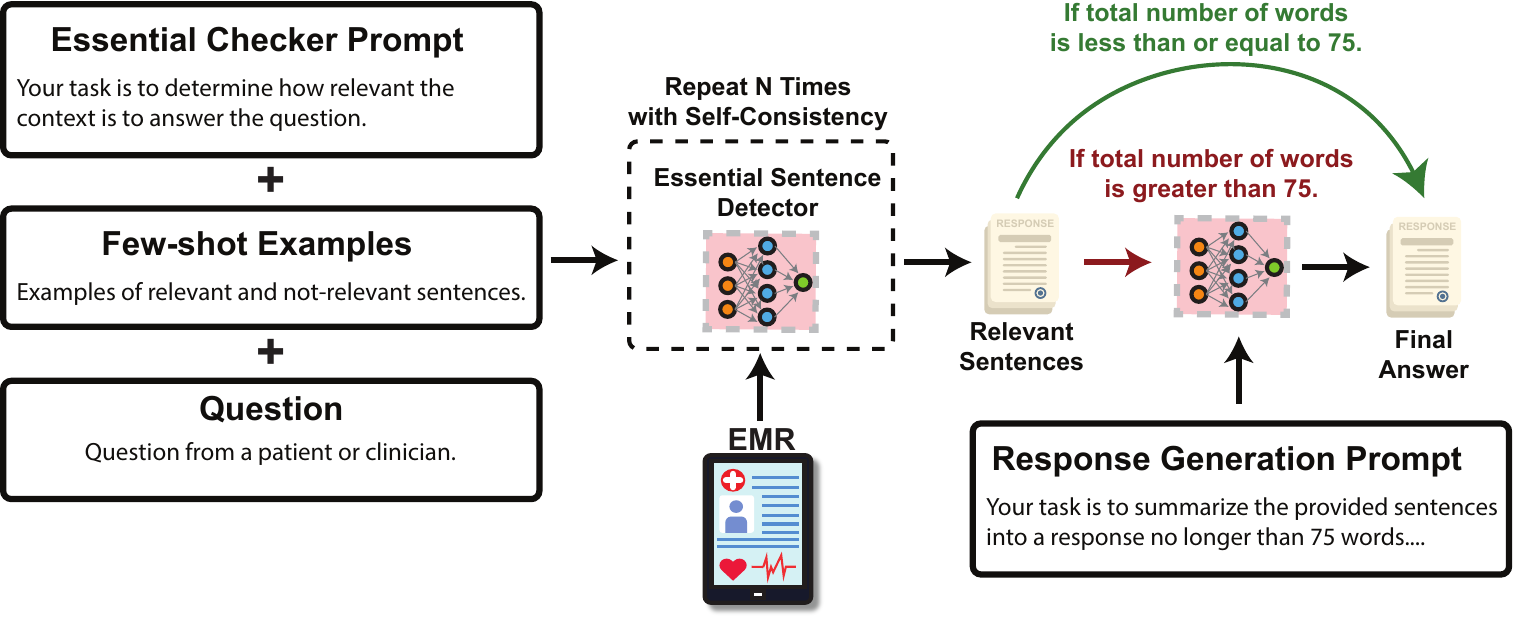}
    \caption{Overview of our multi-step approach for generating clinically grounded answers from electronic health records. An essential sentence detector uses a prompt and few-shot examples to classify sentences from the EMR with self-consistency sampling. Relevant sentences are passed to the response generation step. If the total word count is within the 75-word limit, the sentences are used directly. Otherwise, the sentences are summarized using a language model prompted to stay within the word constraint.}
    \label{fig:overview}
\end{figure*}

\section{Methodology}
Overall, this we aim to generate clinically grounded answers to patient (or clinician) questions using evidence from electronic health records \citep{soni-etal-2025-dataset-patient-needs}.  We provide an overview of our approach in Figure~\ref{fig:overview}. A more complete overview is shown in the Appendix in Figure~\ref{fig:overview2}.
Our approach to the task was two-fold: \textbf{(1.)} we develop an ``relevant sentence identifier'' using few-shot prompting and self-consistency with thresholding.  Specifically, the first objective was to classify each sentence in the note excerpt as \textit{essential},\textit{supplementary}, or \textit{not relevant} to answer the clinician's question\footnote{In initial experiments we found limited difference in performance between using the clinician, patient, or even a combination of clinician an patient questions.}. \textbf{(2.)} we generate final answers to the question using zero-shot prompting that transforms a list of relevant sentences into a 75-word response with citations indicating the sentence used.

Formally, given a natural language question $q$ and a set of context sentences from an electronic health record $C = \{s_1, s_2, \ldots, s_n\}$, the goal is to generate an answer $a$ consisting of at most 75 words, grounded in a subset of the sentences in $C$. Our two-part approach uses large language models (LLMs) for: \textbf{(1)} sentence selection, where we learn a function $f_\text{rel}(q, C) \rightarrow C' \subseteq C$ that identifies relevant sentences, and \textbf{(2}) answer generation, where we use a function $f_\text{gen}(q, C') \rightarrow a$ to compose a fluent, grounded answer. Sentence citations from $C'$ are retained in $a$ to support factual consistency. We provide an overview of our approach in the Appendix.

\subsection{Relevant Sentence Identification}
To identify which sentences in the clinical note are useful for answering the question, we use few-shot prompting with a large language model. Each sentence in the note is independently classified as \textit{essential}, \textit{supplementary}, or \textit{not relevant} to the clinician-formulated question. We construct our few-shot prompts using labeled examples from the development set, where each example includes a question and a sentence with its gold relevance label. A balanced set of 30 such examples (sentences) is randomly sampled and inserted into the prompt for each test case.

\subsection{Self-Consistency and Thresholds}
Overall, we find that the model has a hard time detecting essential (relevant) sentences, and will often default to not-relevant. To mitigate class imbalance issues at inference time, we adopt self-consistency decoding with thresholding. Despite using a pretrained model, the majority class of \textit{not relevant} may be predicted most of the time. Thresholding is used to ensure other classes, especially \textit{essential}, are predicted and can be used for response generation. For each sentence, we sample 20 independent predictions using the LLaMA 8B model with temperature set to 1.0. Thresholds are applied to determine the final label: a sentence is labeled \textit{essential} if it appears in at least 2 out of 20 predictions; if not, it is labeled \textit{supplementary} if it appears at least once; otherwise, it is labeled \textit{not relevant}. This strategy biases the classification toward relevant categories. Increasing the number of samples beyond 20 (e.g., to 111 or 200) yielded limited gains in F1 score on the development set. Prior thresholding attempts used the data description statistics (both median and mean) of the distribution of each classification of sentences per case. For example, the median number of \textit{essential} sentences is 5.5 out of 21 total sentences per case \citep{soni-etal-2025-dataset-patient-needs}. The threshold was set at 26 percent, meaning that of the 20 self-consistency samples, at least 5 would have to be labeled \textit{essential} for the final label to be \textit{essential}. We repeated several attempts with these median and mean thresholds, but these high thresholds yielded F1 scores with several false negatives for the \textit{essential} label.

\begin{table*}[t]
\centering
\resizebox{\linewidth}{!}{%
\begin{tabular}{lrrrrrrrrrrrrrrrrrrrrr}
\toprule
\textbf{Configuration} & \textbf{Ovr} & \textbf{Rel.}  & \textbf{Fact.} & \textbf{SMP} & \textbf{SMR} & \textbf{SMF1} & \textbf{SmP} & \textbf{SmR} & \textbf{SmF1} & \textbf{LMP} & \textbf{LMR} & \textbf{LMF1} & \textbf{LmP} & \textbf{LmR} & \textbf{LmF1} & \textbf{ROU} & \textbf{MED} & \textbf{BE} & \textbf{AS} & \textbf{BL} & \textbf{SA}\\
\midrule
\textsc{8B-Lenient}                 & 42.4 & 32.1 & 52.7 & 44.9 & 75.0 & 51.4 & 41.6 & 71.7 & 52.7 & 61.9 & 73.1 & 60.4 & 54.2 & 71.7 & 60.4 & 20.7 & 32.7 & 27.3 & 30.6 & 1.8 & 59.6 \\
\textsc{8B-Strict}                 & 36.1 & 25.9 & 46.4 & 42.5 & 44.0 & 38.0 & 48.8 & 44.2 & 46.4 & 63.1 & 78.5 & 40.9 & 59.2 & 39.2 & 47.1 & 14.8 & 28.1 & 15.6 & 40.5 & 1.3 & 56.9 \\
\textsc{SharedTask-Baseline}        & 35.9 & 28.7 & 43.1 & 70.3 & 47.1 & 49.4 & 63.4 & 32.6 & 43.1 & 78.5 & 38.9 & 46.5 & 71.8 & 27.0 & 39.2 & 18.7 & 29.4 & 24.2 & 52.1 & 0.2 & 48.0 \\
\midrule

\textsc{GT-Lenient}              & 61.90 & 39.39 & 84.40 & 78.25 & 100 & 85.52 & 73.02 & 100 & 84.40 & 100 & 100 & 100 & 100 & 100 & 100 & 28.52 & 40.62 & 35.43 & 59.21 & 6.61 & 65.95\\
\textsc{GT-Strict}                 & 74.58 & 49.16 & 100 & 100 & 100 & 100 & 100 & 100 & 100 & 100 & 78.25 & 85.52 & 100 & 73.02 & 84.40 & 36.42 & 48.64 & 42.56 & 82.25 & 11.18 & 73.89\\
\midrule
\textsc{8B-Lenient (w/o Thresh.)}             & 29.3 & 24.5 & 34.1 & 47.4 & 29.1 & 30.3 & 49.3 & 26.1 & 34.1 & 58.0 & 28.8 & 31.9 & 57.5 & 22.2 & 32.1 & 15.4 & 22.2 & 13.1 & 36.6 & 2.7 & 57.0 \\
\textsc{8B-Lenient (w/o Thresh. and SC)}      & 26.3 & 22.3 & 30.3 & 34.7 & 25.1 & 24.0 & 43.8 & 23.2 & 30.3 & 42.8 & 25.2 & 25.6 & 50.7 & 19.6 & 28.2 & 10.0 & 17.4 & 11.5 & 30.8 & 2.7 & 57.8 \\
\bottomrule
\end{tabular}}
\caption{Development set performance across configurations. \textbf{Ovr} is the overall score (mean of factuality and relevance). \textbf{Rel.} and \textbf{Fact.} are overall relevance and factuality. \textbf{SMP/SMR/SMF1} and \textbf{LMP/LMR/LMF1} are strict micro/macro precision, recall, and F1 (``essential'' only). \textbf{SmP/SmR/SmF1} and \textbf{LmP/LmR/LmF1} are lenient versions (``essential'' + ``supplementary''). \textbf{ROU}, \textbf{BL}, \textbf{SA}, \textbf{BE}, \textbf{AS}, and \textbf{MED} denote ROUGE-L, BLEU, SARI, BERTScore, AlignScore, and MEDCON.}
\vspace{0em}
\label{tab:overall_scores}
\end{table*}

\subsection{Final Response Generation.}
We use zero-shot prompting for response generation using the Llama 70B quantized model without any in-context examples. Instead, a system prompt (see Appendix) was provided to guide the structure and formatting of the output. For lenient evaluation, both \textit{essential} and \textit{supplementary} sentences were used as relevant sentences to compose the response, while strict evaluation included only sentences labeled as \textit{essential}. Sentences marked as \textit{not relevant} were excluded from all responses. In cases where no \textit{essential} sentences were identified, a placeholder response (``No citations found'') was generated along with a randomly sampled citation (pipe-delimited from 1–10) to satisfy format requirements that sentences must have at least one citation. If the extracted content was under the 75-word limit, the sentence set was used directly without further processing. We did not attempt to pass these sentences of less than 75 words to the model to make the response even more concise because we were concerned that further reduction might negatively impact ROUGE and BERT scores by decreasing n-gram overlap. Only when the combined sentence content exceeded the limit, we used the Llama 70B model to summarize the selected sentences into a coherent answer under the constraint with the zero-shot prompt. If citations were removed by the Llama model, citations for any \textit{essential} or \textit{supplementary} sentences excluded due to length were appended to the final sentence of the response.

\subsection{Models.}
We evaluated both LLaMA 3.1 8B and 3.1 70B (w416b quantization) models~\cite{grattafiori2024llama} on the development set for sentence classification and answer generation.\footnote{We also evaluated UltraMedical~\cite{zhang2024ultramedical}, but it did not outperform the LLaMA models in initial experiments. Although UltraMedical showed some promise, its responses were consistently too lengthy and overly elaborate, making it unsuitable for the 75-word constraint on the test set.} Based on development performance, we selected two configurations for test set evaluation: \textbf{(1)} a entirely 70B pipeline for both sentence selection and answer generation (denoted \textsc{70B-Lenient} and \textsc{70B-Strict}), and \textbf{(2)} a hybrid configuration using LLaMA 8B for essential sentence identification and LLaMA 70B for response generation called  \textsc{8B-Lenient} and  \textsc{8B-Strict}.

\section{Results}
Here, we present our results from several attempts using the development data set and our three submissions to the competition using the test data set.

\subsection{Evaluation Metrics}
The ArchEHR scoring script was used to evaluate all attempts for classification accuracy (strict and lenient F1 scores) and response quality, including fluency, relevance, and medical accuracy, using metrics such as BLEU~\cite{papineni2002bleu}, ROUGE~\cite{lin2004rouge}, SARI~\cite{xu2016optimizing}, BERTScore~\cite{zhang2019bertscore}, MedCon~\cite{yim2023aci}, and AlignScore~\cite{zha2023alignscore}. The scoring script parsed the responses for citations and then, if needed, truncated the response to 75 words. Only the 75 words and the pipe-delimited citations were evaluated.

\subsection{Validation Results.}  
We report the results on the dev set using leave-one-out cross-validation in Table~\ref{tab:overall_scores}. The best performance on the development set was achieved using the \textsc{LLaMA3-8B} model to identify both \textit{essential} and \textit{supplementary} sentences, followed by response generation using the \textsc{LLaMA3-70B} model (denoted \textsc{Lenient (8B)}). This configuration outperformed all others, including strict variants that relied solely on \textit{essential} sentences.

To assess the potential upper bound of the task, we evaluate ground truth (GT) configurations that assume perfect classification. \textsc{GT-Strict} uses only the gold \textit{essential} sentences, while \textsc{GT-Lenient} includes both gold \textit{essential} and \textit{supplementary} sentences as input to the response generation model. These oracle settings achieved substantially higher scores, 74.58 and 61.90, respectively, demonstrating the significant headroom remaining for improving sentence selection models. This comparison also highlights the performance of our lenient model, which achieves 42.37 overall.

Finally, we conducted ablations to examine the effect of self-consistency and thresholding in the classification process. Removing thresholding alone reduced overall performance to 26.34, while removing both thresholding and self-consistency resulted in a score of 29.31 (Table~\ref{tab:overall_scores}). These declines were primarily driven by reduced F1 scores for sentence classification, underscoring the importance of threshold-based calibration in achieving stable, high-quality predictions.

\subsection{Competition Results.} 
The 8B model was used to classify sentences using lenient evaluation metrics, and the 70B model was then used to generate responses based on those classifications. This combination (``8B-lenient'') outperformed the 70B model when it was used alone to both classify sentences and generate responses (``70B-lenient'' and ``70B-strict''; Table~\ref{tab:overall_scores_comp} in the Appendix). 
For classification tasks, the 8B model had better consistency in label prediction and produced more factually correct and relevant answers than the 70B quantized models. Furthermore, the balanced recall and precision scores indicate that the thresholds were well-established as the model was able to identify most of the essential sentences. The increased performance in sentence classification led to higher-quality response generation and improved the response generation metrics. Responses had better alignment, quality, and included more medical-specific content. Despite having fewer parameters, the 8B model outperformed the 70B quantized model across almost every metric, especially in classification, which showed to be a key point in generating high-quality responses.

\begin{table}[t]
\centering
\resizebox{.9\linewidth}{!}{%
\begin{tabular}{lcccc}
\toprule
\textbf{Class} & \textbf{TP} & \textbf{FP} & \textbf{FN} & \textbf{TN} \\
\midrule
\textbf{Essential}     & 64  & 67  & 74  & 223 \\
\textbf{Supplementary} & 18  & 89  & 33  & 288 \\
\textbf{Not-relevant}  & 130 & 60  & 109 & 129 \\
\bottomrule
\end{tabular}}
\caption{Confusion Matrix for ``Strict'' Results.}\vspace{0em}
\label{tab:confusion_matrix_strict}
\end{table}

\begin{comment}
\begin{table}[ht]
\centering
\begin{tabular}{|l|c|c|c|c|}
\hline
\textbf{Class} & \textbf{TP} & \textbf{FP} & \textbf{FN} & \textbf{TN} \\
\hline
Essential     & 64  & 67  & 74  & 223 \\
Supplementary & 18  & 89  & 33  & 288 \\
Not-relevant  & 130 & 60  & 109 & 129 \\
\hline
\end{tabular}
\caption{Confusion Matrix for "Strict" Classification Predictions}
\label{tab:confusion_matrix}
\end{table}
\end{comment}

\subsection{Error Analysis.}
While the model shows the ability to differentiate between classes, performance was negatively affected by class imbalance. The overwhelming number of \textit{not relevant} sentences and the relatively small number of \textit{supplementary} sentences led to label mismatches and reduced classification accuracy.

In the strict classification setting, the model was expected to predict three distinct classes. As shown in Table~\ref{tab:confusion_matrix_strict}, the \textit{supplementary} class proved particularly difficult to identify, with only 18 out of 51 instances correctly predicted. High false negative rates for both \textit{essential} and \textit{not relevant} sentences suggest that important information was often missed, and irrelevant content was not reliably excluded.

In the lenient setting, where \textit{essential} and \textit{supplementary} sentences were grouped into a single class, the task was reduced to binary classification (Table~\ref{tab:confusion_matrix_binary_len}). This improved recall for relevant content, and the model successfully identified a larger number of relevant sentences. However, the distinction between \textit{essential} and \textit{supplementary} information introduced ambiguity. While the lenient setup benefited answer generation on the development set, it also produced a high number of false positives, likely due to the low thresholding strategy that aimed to capture as many relevant sentences as possible. 

To better understand these trends, we conducted a manual error analysis on development set predictions. One common error involved \textit{not relevant} sentences being misclassified as \textit{essential} or \textit{supplementary}. For example, in response to the clinician question ``Why was a procedure used instead of a medication?'', two sentences containing only the acronym of the procedure (which also appeared in the question) were incorrectly labeled \textit{essential}. Although these sentences referenced the procedure, they did not explain the reasoning behind it. This suggests that the model may rely too heavily on lexical overlap without considering the deeper intent of the question.

We also observed the opposite error, where sentences labeled as \textit{essential} were misclassified as \textit{not relevant}. In one case, the clinician question concerned a patient's oxygen flow, and a relevant sentence referenced ``hypoxia'' and ``respiratory failure''. These terms are clinically important for evaluating oxygen status, yet the model failed to recognize the connection. This misclassification may be due to the model's reliance on surface features rather than contextual relationships.

\begin{table}[t]
\centering
\resizebox{.9\linewidth}{!}{%
\begin{tabular}{lrrrr}
\toprule
\textbf{Class} & \textbf{TP} & \textbf{FP} & \textbf{FN} & \textbf{TN} \\
\midrule
\textbf{Essential}     & 129 & 109 & 60  & 130 \\
\textbf{Not-relevant}  & 130 & 60  & 109 & 129 \\
\bottomrule
\end{tabular}}
\caption{Confusion matrix for ``Lenient'' Results.}\vspace{0em}
\label{tab:confusion_matrix_binary_len}
\end{table}

Ambiguity in the \textit{supplementary} label also introduced challenges. In one example, a two-part clinician question asked about the lasting effects of poisoning and the patient's confusion. The model often misclassified \textit{essential} sentences as \textit{supplementary} or vice versa, suggesting it struggled to distinguish between past clinician explanations and future clinical concerns. Additionally, a sentence mentioning psychiatry was misclassified as \textit{not-relevant} instead of \textit{supplementary}, likely because the model failed to connect psychiatric care with the patient's mental state in the question.

\section{Conclusion}
Our approach to the ArchEHR-QA Shared Task showed that sentence classification is essential for generating high-quality, grounded responses from electronic health records. Using few-shot prompting with self-consistency and thresholding improved performance, and the smaller \textsc{LLaMA3.1-8B} model outperformed the larger 70B model in identifying relevant sentences. However, distinguishing supplementary content remained difficult due to label imbalance.

Future work should explore incorporating sentence context and document structure to improve classification, along with adaptive thresholding based on model confidence. Fine-tuning with clinician feedback and expanding evaluation to include human judgments will be important for improving real-world reliability and clinical applicability.

\section*{Acknowledgments}
This material is based upon work supported by the National Science Foundation (NSF) under Grant~No. 2145357.

\newpage
\section*{Limitations}
This work has several limitations. First, the dataset was relatively small, consisting of only 20 development cases and 100 test cases, which may limit the generalizability of the results. Additionally, the evaluation relied solely on quantitative metrics, without manual review of patient context and medical accuracy. It also lacked evaluation of personable aspects such as empathy and professionalism. Finally, the imposed word limit on responses introduced a scoring bias, particularly disadvantaging longer or complex patient cases that required more nuanced explanations.

Another limitation lies in the reliance on self-consistency thresholding as a heuristic rather than a learned calibration method. Although it improved performance, the threshold values were tuned manually and may not generalize well across datasets with different distributions of relevance labels. Future work could explore adaptive or data-driven methods to calibrate sentence selection confidence.

Additionally, while the 8B model outperformed the 70B model in sentence classification, this may reflect the effects of quantization, prompt format sensitivity, or differences in instruction tuning. These variables were not systematically controlled or analyzed. Further investigation is needed to isolate whether smaller models offer consistent advantages or whether specific tuning strategies are responsible for the performance gains.

The current approach treats each sentence independently during classification, ignoring the surrounding context that may be critical in understanding clinical relevance. Sentences referring to previous or subsequent medical events could be misclassified due to this lack of discourse awareness. Integrating document-level context or sequential modeling could help mitigate this issue.

\bibliographystyle{acl_natbib}
\bibliography{custom}

\appendix
\section{Appendix}
\label{sec:appendix}

\begin{table*}[t]
\centering
\resizebox{\linewidth}{!}{%
\begin{tabular}{lrrrrrrrrrrrrrrrrrrrrr}
\toprule
\textbf{Configuration} & \textbf{Overall} & \textbf{Rel.}  & \textbf{Fact.} & \textbf{SMP} & \textbf{SMR} & \textbf{SMF1} & \textbf{SmP} & \textbf{SmR} & \textbf{SmF1} & \textbf{LMP} & \textbf{LMR} & \textbf{LMF1} & \textbf{LmP} & \textbf{LmR} & \textbf{LmF1} & \textbf{R} & \textbf{Med} & \textbf{BE} & \textbf{AS} & \textbf{BL} & \textbf{SA}\\
\midrule
8B Results            & 40.45 & 27.92 & 52.97 & 45.06 & 77.3  & 52.59 & 43.71 & 67.22 & 52.97 & 49.6  & 77.43 & 56.65 & 47.02 & 68.39 & 55.73 & 11.11 & 17.75 & 29.4  & 56.58 & 0.72 & 22.7  \\
Organizers   Baseline                     & 30.7  & 27.8  & 33.6  & 77.4  & 31.5  & 39    & 71.6  & 21.9  & 33.6  & 83    & 30.8  & 39.9  & 77    & 22.3  & 34.6  & 15.2  & 25.6  & 20.5  & 57.7  & 0.1  & 47.8  \\
70B   Lenient Results & 29.93 & 18.59 & 41.28 & 44.4  & 51.06 & 41.7  & 38.67 & 44.26 & 41.28 & 47.41 & 50.71 & 43.64 & 41.42 & 44.83 & 43.06 & 8.83  & 12.56 & 10.78 & 54.81 & 1.42 & 12.86 \\
70B Strict   Results  & 28.05 & 18.92 & 37.18 & 43.77 & 44.53 & 38.3  & 36.27 & 38.15 & 37.18 & 46.88 & 44.4  & 39.97 & 38.64 & 38.44 & 38.54 & 8.77  & 12.32 & 11.33 & 54.96 & 1.3  & 12.71 \\
\bottomrule
\end{tabular}}
\caption{Overall Score Comparison of Model Configurations for Test Data.  \textbf{Ovr} is the overall score (mean of factuality and relevance). \textbf{Rel.} and \textbf{Fact.} are overall relevance and factuality. \textbf{SMP/SMR/SMF1} and \textbf{LMP/LMR/LMF1} are strict micro/macro precision, recall, and F1 (``essential'' only). \textbf{SmP/SmR/SmF1} and \textbf{LmP/LmR/LmF1} are lenient versions (``essential'' + ``supplementary''). \textbf{ROU}, \textbf{BL}, \textbf{SA}, \textbf{BE}, \textbf{AS}, and \textbf{MED} denote ROUGE-L, BLEU, SARI, BERTScore, AlignScore, and MEDCON.}
\label{tab:overall_scores_comp}
\end{table*}

\begin{figure*}
    \centering
    \includegraphics[width=\linewidth]{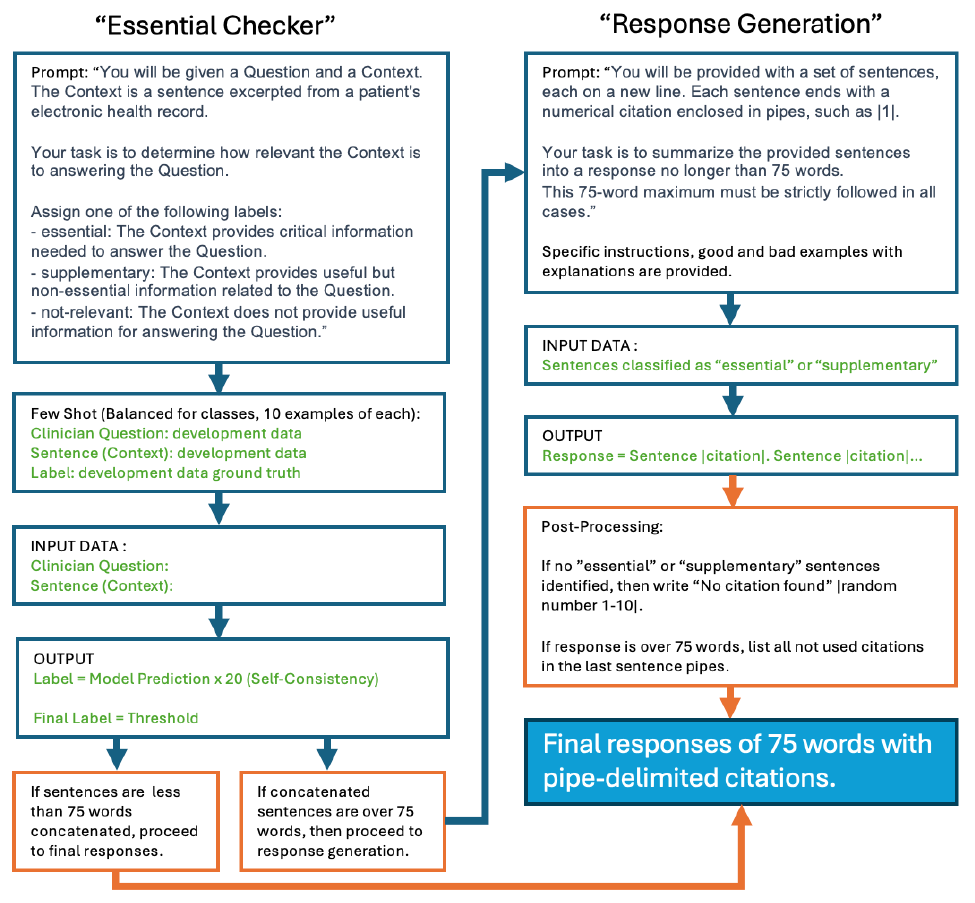}
    \caption{Overall Method Figure.}
    \label{fig:overview2}
\end{figure*}

\subsection*{A. Prompt for Sentence Relevance Classification}

We use a single, structured prompt to classify the relevance of EHR sentences with respect to a clinical question. For clarity, we display the full prompt below in a formatted box, broken into the system instruction and examples.

\begin{tcolorbox}[colback=gray!5!white, colframe=black, fontupper=\small, width=\linewidth, boxsep=4pt, title=Prompt for ``Essential Checker'']
"You will be given a Question and a Context. The Context is a sentence excerpted from a patient's electronic health record.\\

\vspace{0.5em}
Your task is to determine how relevant the Context is to answering the Question.\\

\vspace{0.5em}
Assign one of the following labels:\\
- essential: The Context provides critical information needed to answer the Question.\\
- supplementary: The Context provides useful but non-essential information related to the Question.\\
- not-relevant: The Context does not provide useful information for answering the Question.\\

\vspace{0.5em}
Important: Output only the label — "essential", "supplementary", or "not-relevant". Do not include any other text.
\end{tcolorbox}

\vspace{0.5em}
\noindent \textbf{Examples:}\footnote{Examples have been changed to ensure anonyminity of data.}

\begin{tcolorbox}[colback=white, colframe=black, fontupper=\small, width=\linewidth, boxsep=4pt] 

Question: What medications is the patient currently taking?\\
Context: The patient is currently prescribed metformin and lisinopril.\\
Label: essential

\vspace{0.5em}
Question: Has the patient experienced any recent falls?\\
Context: The patient reports no falls over the past six months.\\
Label: essential

\vspace{0.5em}
Question: What is the patient’s preferred pharmacy?\\
Context: The patient prefers CVS Pharmacy on Main Street.\\
Label: essential

\vspace{0.5em}
Question: What medications is the patient currently taking?\\
Context: The patient lives with their daughter and two grandchildren.\\
Label: not-relevant

\vspace{0.5em}
Question: Has the patient experienced any recent falls?\\
Context: The patient has a history of osteoarthritis in the knees.\\
Label: supplementary

\vspace{0.5em}
Question: What is the patient’s preferred pharmacy?\\
Context: The patient reports good control of their blood sugar levels.\\
Label: not-relevant
\end{tcolorbox}

\vspace{0.5em}
\noindent \textbf{Additional Examples:}

\begin{tcolorbox}[colback=white, colframe=black, fontupper=\small, width=\linewidth, boxsep=4pt]

Question: What medications is the patient currently taking?\\
Context: The patient reports an allergy to penicillin.\\
Label: supplementary

\vspace{0.5em}
Question: Has the patient experienced any recent falls?\\
Context: Patient noted to have unsteady gait and occasional dizziness.\\
Label: supplementary

\vspace{0.5em}
Question: What is the patient’s preferred pharmacy?\\
Context: The patient was discharged home with follow-up scheduled in two weeks.\\
Label: not-relevant

\vspace{0.5em}
Question: What medications is the patient currently taking?\\
Context: At discharge, the patient was advised to continue taking atorvastatin daily.\\
Label: essential

\vspace{0.5em}
Question: Has the patient experienced any recent falls?\\
Context: The patient was admitted after slipping on ice and fracturing their wrist last month.\\
Label: essential
\end{tcolorbox}

\subsection*{B. Prompt for Answer Generation}

We use a single zero-shot prompt to guide answer generation. The model receives a list of pre-selected sentences with citations and is asked to generate a 75-word summary with citation formatting preserved.

\begin{tcolorbox}[colback=gray!5!white, colframe=black, fontupper=\small, width=\linewidth, boxsep=4pt, title=Prompt for ``Response Generator'']
"You will be provided with a set of sentences, each on a new line. Each sentence ends with a numerical citation enclosed in pipes, such as |1|.\\

\vspace{0.5em}
Your task is to summarize the provided sentences into a response no longer than 75 words.\\
This 75-word maximum must be strictly followed in all cases.\\

\vspace{0.5em}
Do not add any notes, comments, or additional text after the summary.\\
This will result in the response exceeding the 75-word limit.\\

\vspace{0.5em}
Each sentence in your output should start on a new line.\\
Each sentence must have one or more citations at the end, formatted as integers inside pipes (e.g., |2| or |3,5,7|).\\

\vspace{0.5em}
When combining multiple original sentences into one, list all relevant citations in order, separated by commas inside a single pair of pipes (e.g., |2,4,5|).\\
If multiple sequential citations are combined, list them individually, not as a range (e.g., |7,8,9,10|, not |7-10|).\\

\vspace{0.5em}
If any sentences from the input are omitted completely from your summary, their citations must still be preserved by adding them to the final sentence’s citation list.\\

\vspace{0.5em}
Only output the summarized response. Do not include any commentary, labels, or additional text."
\end{tcolorbox}

\noindent \textbf{Examples:}

\begin{tcolorbox}[colback=white, colframe=black, fontupper=\small, width=\linewidth, boxsep=4pt, title=Example 1 — Input and Output]

\textbf{Input:}\\
The company launched a new product in April |1|.\\
Sales exceeded expectations within the first month |2|.\\
Customer feedback highlighted a few technical issues |3|.\\
The technical team promised a software update to address concerns |4|.

\vspace{0.5em}
\textbf{Output:}\\
The company launched a new product in April, and sales exceeded expectations in the first month |1,2|.\\
Customer feedback highlighted technical issues, and the technical team promised a software update to address them |3,4|.
\end{tcolorbox}

\begin{tcolorbox}[colback=white, colframe=black, fontupper=\small, width=\linewidth, boxsep=4pt, title=Example 2 — Output]
A new downtown cafe offering organic food received praise for its atmosphere but some criticism for high prices |1,2,3,4|.\\
It plans to expand to a second location next year |5|.
\end{tcolorbox}

\begin{tcolorbox}[colback=white, colframe=black, fontupper=\small, width=\linewidth, boxsep=4pt, title=Example 3 — Output]
The software update brought a redesigned interface and improved navigation |1,2|.\\
Although users reported new bugs, a patch issued two weeks later resolved major issues but caused minor compatibility problems on older devices |3,4,5|.
\end{tcolorbox}

\begin{tcolorbox}[colback=white, colframe=black, fontupper=\small, width=\linewidth, boxsep=4pt, title=Bad Example (What \textbf{Not} to Do)]

\textbf{Input:}\\
The research team published their findings in a leading journal |1|.\\
They discovered a new species of bacteria in the Arctic |2|.\\
The bacteria showed resistance to extreme cold temperatures |3|.\\
Further studies are needed to understand its potential applications |4|.

\vspace{0.5em}
\textbf{Output (Incorrect):}\\
The research team published their findings about a new cold-resistant bacteria discovered in the Arctic |1-3|.\\
Further studies are needed to understand its applications |4|.

\vspace{0.5em}
\textbf{Issues:}
\begin{itemize}
    \item Incorrect citation format: |1-3| is a range, but it should be |1,2,3|.
    \item Word count and sentence coverage are fine, but citation formatting makes this output invalid.
\end{itemize}
\end{tcolorbox}

\end{document}